\def\year{2019}\relax
\documentclass[letterpaper]{article} %DO NOT CHANGE THIS
\usepackage{aaai19}   %Required
\usepackage{times}    %Required
\usepackage{helvet}   %Required
\usepackage{courier}  %Required
\usepackage{url}      %Required
\usepackage{graphicx} %Required
\usepackage{amsmath,amssymb}
\usepackage{booktabs}
\usepackage{multirow}
\usepackage{subcaption}
\usepackage{xcolor}
\usepackage[export]{adjustbox}

\usepackage[pagebackref=false,breaklinks=true,letterpaper=true,colorlinks,urlcolor=blue,citecolor=blue,linkcolor=blue,bookmarks=false]{hyperref}

\DeclareMathAlphabet      {\mathbfit}{OML}{cmm}{b}{it}

\newcommand{\ra}[1]{\renewcommand{\arraystretch}{#1}}

\newcommand{\xmark}{\ding{55}}
\usepackage{pifont}

\newcommand{\etal}{\textit{et~al}\mbox{.}}
\newcommand{\eg}{e.g.,\ }
\newcommand{\ie}{i.e.,\ }
\newcommand{\bbR}{{\mathbb{R}}}

\newlength{\twoimg}

\newlength\paramargin
\newlength\figmargin
\newlength\tablemargin
\newlength\secmargin
\newlength\figcapmargin
\newlength\tablecapmargin

\begin{document}
\title{Learning Resolution-Invariant Deep Representations for Person Re-Identification}
% * <b03901148@ntu.edu.tw> 2018-08-27T13:13:05.037Z:
%
% ^.
\author{Yun-Chun Chen\textsuperscript{1}\thanks{ indicates equal contributions.}, Yu-Jhe Li\textsuperscript{1,2}\footnotemark[1], Xiaofei Du\textsuperscript{3}, and Yu-Chiang Frank Wang\textsuperscript{1,2}\\
\textsuperscript{1}\hspace{1pt}Department of Electrical Engineering, National Taiwan University\\
%\textsuperscript{2}\hspace{1pt}Graduate Institute of Communication Engineering, National Taiwan University, Taiwan\\
\textsuperscript{2}\hspace{1pt}MOST Joint Research Center for AI Technology and All Vista Healthcare\\
\textsuperscript{3}\hspace{1pt}Umbo Computer Vision\\
Email: \{b03901148, r06942074, ycwang\}@ntu.edu.tw, xiaofei.du@umbocv.com
% Association for the Advancement of Artificial Intelligence\\
% 2275 East Bayshore Road, Suite 160\\
% Palo Alto, California 94303\\
}
\maketitle

\begin{abstract}
Person re-identification (re-ID) solves the task of matching images across cameras and is among the research topics in vision community. Since query images in real-world scenarios might suffer from resolution loss, how to solve the resolution mismatch problem during person re-ID becomes a practical problem. Instead of applying separate image super-resolution models, we propose a novel network architecture of Resolution Adaptation and re-Identification Network (RAIN) to solve cross-resolution person re-ID. Advancing the strategy of adversarial learning, we aim at extracting resolution-invariant representations for re-ID, while the proposed model is learned in an end-to-end training fashion. Our experiments confirm that the use of our model can recognize low-resolution query images, even if the resolution is not seen during training. Moreover, the extension of our model for semi-supervised re-ID further confirms the scalability of our proposed method for real-world scenarios and applications.
\end{abstract}

\section{Introduction}

Aiming at matching images of the same person across different camera views, person re-identification (re-ID)~\cite{zheng2016person} is among the active research topics in computer vision and machine learning. With a wide range of applications ranging from video surveillance to computational forensics, person re-ID has received substantial attention of communities from both academia and industry. Nevertheless, with the presence of background clutters, viewpoint and pose changes, and even occlusion, person re-ID remains a very challenging task.

\begin{figure}[t]
  \begin{center}
  \includegraphics[width=0.48\textwidth]{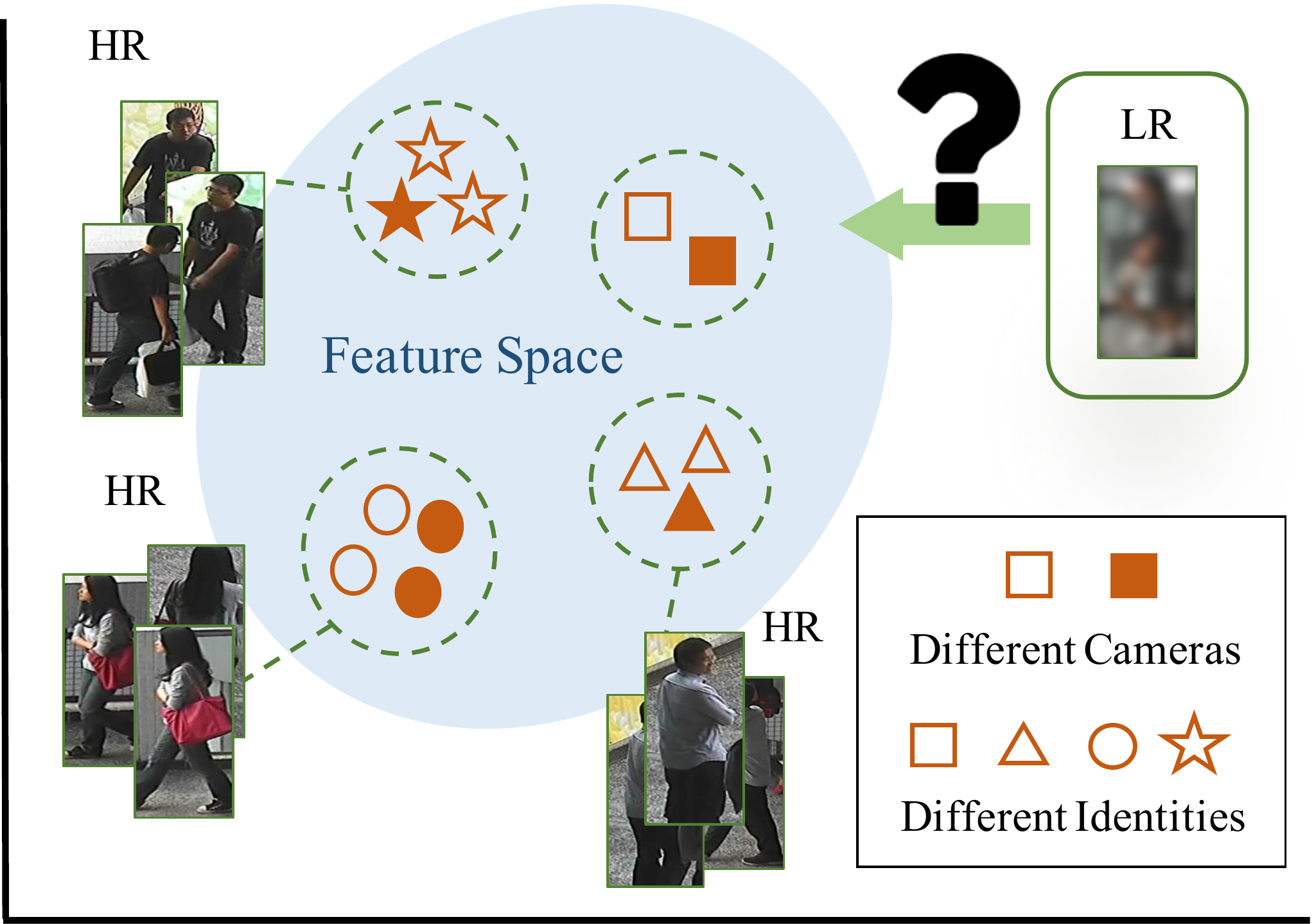}
  \caption{Illustration and challenges of cross-resolution person re-identification (re-ID). In addition to recognizing images across different camera views, one also needs to match cross-resolution images.}
  \label{fig:teaser}
  \end{center}
\end{figure}

While a number of methods~\cite{lin2017improving,hermans2017defense,zhong2017camera,si2018dual} have been proposed to address the aforementioned issues in person re-ID, these methods typically assume that the images (both gallery and query) are of similar or sufficient resolution. However, this assumption may not hold in real-world scenarios, since image resolution may vary drastically due to the distance between the camera and the person of interest. For instance, images captured by surveillance cameras (i.e., the queries to be recognized) are often of low resolution (LR) whereas the gallery ones typically have high resolution (HR). However, directly matching an LR query image against the HR gallery ones would entail a non-trivial \emph{resolution mismatch} problem, as illustrated in Figure~\ref{fig:teaser}.

To address cross-resolution person re-ID, one can simply up-sample the LR images by leveraging super-resolution (SR) approaches like~\cite{jiao2018deep,wang2018cascaded} to synthesize HR images. However, since these two tasks are addressed separately, there is no guarantee that synthesized HR outputs would result in satisfactory re-ID performances. Moreover, if the input image resolution is not seen by the SR model, then one cannot properly recover the HR outputs. Later in the experiments, we will verify the above issues.

In this paper, we propose a novel \emph{Resolution Adaptation and re-Identification Network (RAIN)} for cross-resolution person re-ID. Based on the generative adversarial network (GAN)~\cite{goodfellow2014generative} with an \textit{end-to-end learning} strategy, our RAIN is trained to extract \textit{resolution-invariant} image representations, without the limitation (or assumption) of the use of LR inputs with pre-determined resolutions. More specifically, our RAIN is able to handle unseen LR images with satisfactory re-ID performances. For example, given training LR images with $128$ $\times$ $128$ pixels and $256$ $\times$ $256$ pixels, our model is able to recognize query images with $64$ $\times$ $64$ pixels as well (prior re-ID methods requiring SR models may not properly handle LR images with unseen resolution). Finally, as image labeling is of high labor cost in real-world applications, we conduct a series of \emph{semi-supervised} experiments, which supports the use and extension of our RAIN for cross-resolution person re-ID in such practical yet challenging settings.

The contributions of this paper are highlighted below:
\begin{itemize}
  \item We present an end-to-end trainable network that learns resolution-invariant deep representations for cross-resolution person re-ID.
  \item Our advance multi-level adversarial network components in our proposed architecture effectively aligns and extracts feature representations across resolutions.
  \item We demonstrate the robustness of our model in handling a range of (and even unseen) resolutions for LR query inputs, while standard SR models are required to train on images with particular resolutions.
  \item Extensive experiments are performed to verify the effectiveness of our model, and confirm its use for re-ID in semi-supervised settings.
\end{itemize}

\section{Related Work}

Person re-ID has been widely studied in the literature. Most of the existing methods~\cite{cheng2016person,lin2017improving,kalayeh2018human,si2018dual,chang2018multi,li2018harmonious,liu2018pose,wei2018person,song2018mask,chen2018group,shen2018deep} focus on tackling the challenges of matching images with viewpoint and pose variations, or those with background clutter or occlusion presented. For example, Liu~\etal~\cite{liu2018pose} develop a pose-transferable GAN-based~\cite{goodfellow2014generative} framework to address image pose variations. Chen~\etal~\cite{chen2018group} integrate the conditional random field (CRF) with deep neural networks to learn more consistent multi-scale similarity metrics. The DaRe~\cite{wang2018resource} combines the feature embeddings extracted from different convolutional layers into a single embedding to train the model in a supervised fashion. Several attention-based methods~\cite{si2018dual,li2018harmonious,song2018mask} are further proposed to focus on learning the discriminative parts to mitigate the effect of background clutter. While promising results have been presented, the above approaches typically assume that all images (both query and gallery) are of the same (or similar) resolution, which might not be practical in real-world re-ID applications.

To address the challenging resolution mismatch problem, a couple of methods~\cite{li2015multi,jing2015super,wang2016scale,jiao2018deep,wang2018cascaded,CADNet} have been recently proposed. Li~\etal~\cite{li2015multi} present a joint learning framework that simultaneously optimizes cross-scale image domain alignment and discriminant distance metric modeling. The SLD$^2$L~\cite{jing2015super} learns a pair of HR and LR dictionaries and the mapping between the feature representations of HR and LR images. Wang~\etal~\cite{wang2016scale} explore the scale-distance function space by varying the image scale of LR images when matching with HR ones. Nevertheless, the above methods employ hand-crafted descriptors, which might limit the generalization of their re-ID capability.

Driven by the recent success of convolutional neural networks (CNNs), a few CNN-based re-ID methods~\cite{jiao2018deep,wang2018cascaded} are proposed. For example, the SING~\cite{jiao2018deep} comprises an SR network and a person re-ID model to address the LR re-ID problem. Wang~\etal~\cite{wang2018cascaded} propose the CSR-GAN which cascades multiple SR-GANs~\cite{ledig2017photo} in series to alleviate the resolution mismatch problem. Although remarkable improvements have been presented, the aforementioned methods require the learning of a separate SR model. Treating SR and re-ID as independent tasks, there is no guarantee that solving one task well would be preferable for addressing the other. Moreover, if the resolution of the LR query input is not seen during training, the CSR-GAN~\cite{wang2018cascaded} cannot directly apply the learned SR models for synthesizing the HR images whereas the SING~\cite{jiao2018deep} requires to fuse the results produced by multiple learned models, each of which is specifically designed for a particular resolution. Namely, such models cannot be easily extended to cross-resolution person re-ID.

To overcome the above limitations, our method advances the architecture of the GAN and the autoencoder, which learns cross-resolution deep image representations for re-ID purposes. Our method not only allows LR queries with unseen resolution, but can be extended for solving cross-resolution re-ID in semi-supervised settings. The details of our proposed model will be discussed in the next section.

\section{Proposed Method}

\begin{figure*}[t]
  \centering
  \includegraphics[width=.9\linewidth]{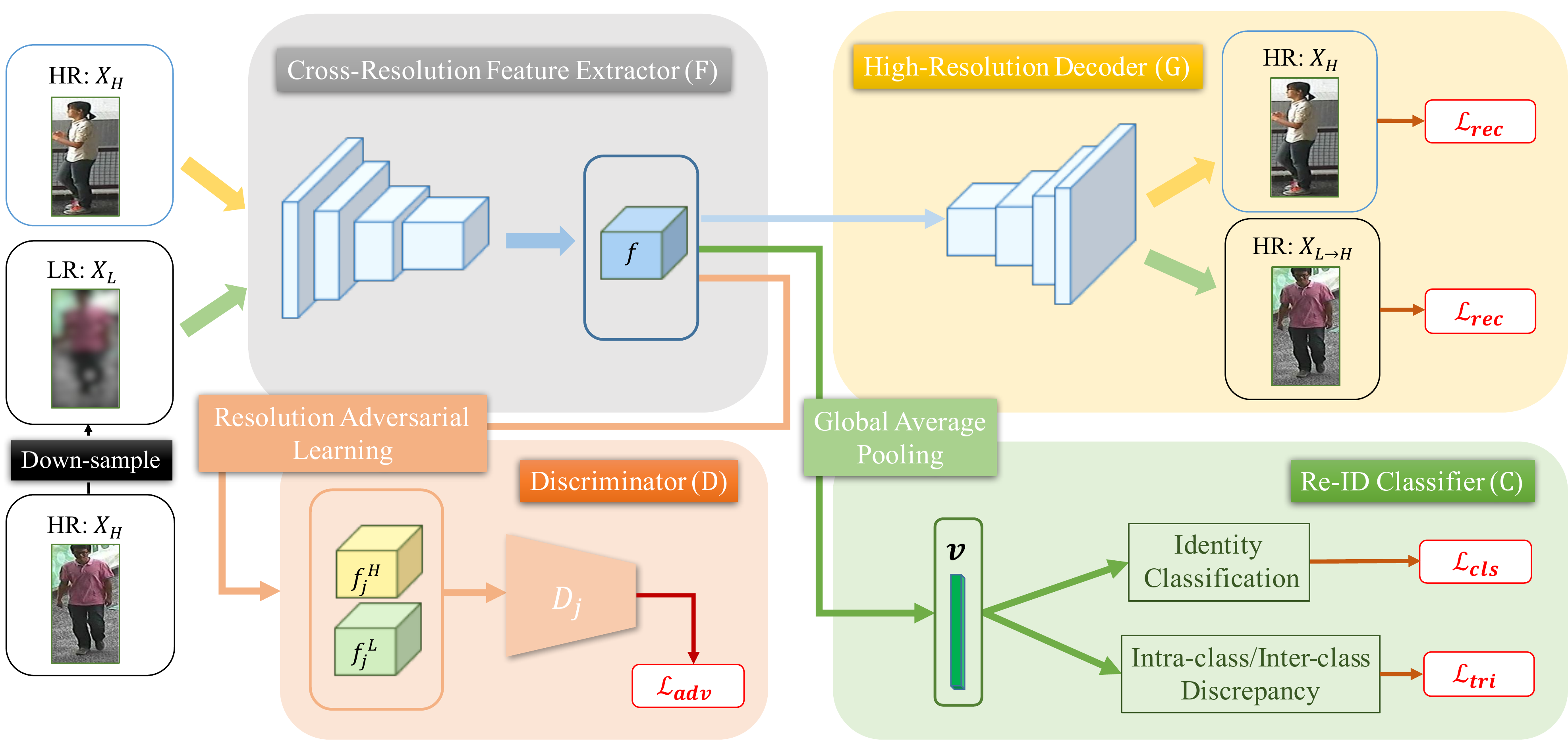}
  \caption{\textbf{Overview of the proposed Resolution Adaptation and re-Identification Network (RAIN)}. The RAIN consists of a cross-resolution feature extractor $\mathcal{F}$ (in gray), a high-resolution decoder $\mathcal{G}$ (in yellow), a resolution discriminator $\mathcal{D}$ (in orange), and a re-ID classifier $\mathcal{C}$ (in green). The associated loss functions (in white) are the high-resolution reconstruction loss $\mathcal{L}_\mathrm{rec}$, adversarial loss $\mathcal{L}_\mathrm{adv}$, classification loss $\mathcal{L}_\mathrm{cls}$, and the triplet loss $\mathcal{L}_\mathrm{tri}$. Note that $j$ denotes the index of feature level.}
  \label{fig:Model}
\end{figure*}

\subsection{Notations and Algorithmic Overview}

For the sake of completeness, we first define the notations to be used in this paper. We assume that we have access to a set of $N$ HR images $X_H = \{x_i^H\}_{i=1}^N$ with the associated label set $Y_H = \{y_i^H\}_{i=1}^N$, where $x_i^H \in \bbR^{H \times W \times 3}$ and $y_i^H \in \bbR$ represent the $i^{th}$ HR image and its corresponding identity label, respectively. To synthesize LR images for training purposes, we generate a \emph{synthetic} image set $X_L = \{x_i^L\}_{i=1}^N$ by down-sampling each image in $X_H$, followed by resizing them back to the original image size via bilinear up-sampling (\ie $x_i^L \in \bbR^{H \times W \times 3}$), where $x_i^L$ is the synthetic LR image associated with $x_i^H$ (with same label). Thus, the label set $Y_L$ for $X_L$ is identical to $Y_H$.

To achieve cross-resolution person re-ID, we present an end-to-end trainable network, \emph{Resolution Adaptation and re-Identification Network (RAIN)}. As presented in Figure~\ref{fig:Model}, our RAIN learns resolution-invariant deep representations from training HR and LR images (note that we only need to down-sample the HR training images to produce the LR ones).

As for testing, our proposed RAIN allows query images with varying resolutions; more specifically, we not only allow query images with HR or LR resolutions which are seen during training, but our model can further handle LR images with intermediate resolutions, or resolutions lower than those of the training images (\ie those not seen during training). In the following subsections, we will detail the network components of RAIN.

\subsection{Architecture of RAIN}
Our proposed network, Resolution Adaptation and re-Identification Network (RAIN), includes a number of network components. The cross-resolution feature extractor $\mathcal{F}$ encodes input images across different resolutions and produces image features for both image recovery and person re-ID. The high-resolution decoder $\mathcal{G}$ reconstructs the encoded cross-resolution features to the HR outputs. The discriminator $\mathcal{D}$ aligns image features across resolutions via adversarial learning, and thus enforces the learning of resolution-invariant features. Finally, the re-ID classifier $\mathcal{C}$ is learned via classification and triplet losses.

{\flushleft {\bf Cross-resolution feature extractor $\mathcal{F}$.}}
Given an HR image $x_H \in X_H$ and an LR image $x_L \in X_L$~\footnote{For simplicity, we would omit the subscript $i$, denote HR and LR images as $x_H$ and $x_L$, and represent their corresponding labels as $y_H$ and $y_L$ in this paper.}, we first forward $x_H$ and $x_L$ to the cross-resolution feature extractor $\mathcal{F}$ to obtain their feature maps. In this paper, we adopt the ResNet-$50$~\cite{he2016deep} as the cross-resolution feature extractor $\mathcal{F}$, which has five residual blocks $\{R_1, R_2, R_3, R_4, R_5\}$. We denote the feature maps extracted from the last activation layer of each residual block as $\{f_1, f_2, f_3, f_4, f_5\}$, where $f_j \in \bbR^{h \times w \times d}$ and $d$ is the number of channels.

Since our goal is to perform cross-resolution person re-ID, we encourage the cross-resolution feature extractor $\mathcal{F}$ to generate similar feature distributions when observing both $X_L$ and $X_H$. To accomplish this, we advance the strategy of adversarial learning, and introduce a discriminator $\mathcal{D}_j$. This discriminator takes in the feature maps $f_j^H$ and $f_j^L$ as inputs to distinguish whether the input feature map is from $X_H$ or $X_L$. Note that $j \in \{1, 2, 3, 4, 5\}$ represents the index of the feature level and $f_j^H$ and $f_j^L$ denote the feature maps of $x_H$ and $x_L$, respectively.

To train the cross-resolution feature extractor $\mathcal{F}$ and the discriminator $\mathcal{D}_j$ with cross-resolution input images $x_H$ and $x_L$, we define the adversarial loss as
\begin{equation}
  \begin{split}
  \mathcal{L}_\mathrm{adv}^{D_j}(X_H, X_L; \mathcal{F}, \mathcal{D}_j) &{}= \mathbb{E}_{x_H \sim X_H}[\log(D_j(f_j^H))] \\
  + &~ \mathbb{E}_{x_L \sim X_L}[\log(1 - D_j(f_j^L))].
  \end{split}
  \label{eq:adv_D_i}
\end{equation}

{\flushleft {\bf High-resolution decoder $\mathcal{G}$.}}
To reduce the information loss in the above feature extraction stage, we introduce a high-resolution decoder $\mathcal{G}$ that takes in the feature map $f_5$ extracted from the cross-resolution feature extractor $\mathcal{F}$ as the input. In contrast to existing autoencoder-based methods that encourage the decoder to recover the original images given the observed latent features (\ie self reconstruction), we explicitly enforce our HR decoder $\mathcal{G}$ to reconstruct the HR images using features derived from the cross-resolution feature extractor $\mathcal{F}$. This would further allow $\mathcal{F}$ to extract cross-resolution image features, while having $\mathcal{G}$ focus on synthesizing the HR outputs.

To achieve the above goal, we impose an HR reconstruction loss $\mathcal{L}_\mathrm{rec}$ between the outputs of the HR decoder $\mathcal{G}$ and the corresponding HR ground truth images. Specifically, the HR reconstruction loss $\mathcal{L}_\mathrm{rec}$ is defined as
\begin{equation}
  \label{eq:rec}
  \begin{split}
  \mathcal{L}_\mathrm{rec}(X_H, X_L; \mathcal{F}, \mathcal{G}) &{}= \mathbb{E}_{x_H \sim X_H}[\|\mathcal{G}(f_5^H) - x_H\|_1] \\
  + &~ \mathbb{E}_{x_L \sim X_L}[\|\mathcal{G}(f_5^L) - x_{L\rightarrow H}\|_1].
  \end{split}
\end{equation}

Note that $x_{L\rightarrow H}$ is the HR image corresponding to $x_L$. Following~\cite{huang2018munit}, we also use the $\ell_1$ norm to calculate the HR reconstruction loss $\mathcal{L}_\mathrm{rec}$, since it would preserve image sharpness.

{\flushleft {\bf Re-ID classifier $\mathcal{C}$.}}
To utilize labeled information of training data for cross-resolution person re-ID, we finally introduce a classifier $\mathcal{C}$ in our RAIN. The input of this classifier is the feature vector $\mathbfit{v}$ from the global average pooling (GAP) layer on the feature map $f_5$, \ie $\mathbfit{v}$ $=$ GAP($f_5$), where $\mathbfit{v} \in \bbR^d$. With person identity as ground truth information, we can compute the negative log-likelihood between the predicted label $\tilde{y} = \mathcal{C}(\mathbfit{v}) \in \bbR^K$ and the ground truth one-hot vector $\hat{y} \in \bbR^K$, and define the classification loss $\mathcal{L}_\mathrm{cls}$ as
\begin{equation}
  \begin{aligned}
  \mathcal{L}_\mathrm{cls}(X_H, X_L; &~ \mathcal{F}, \mathcal{C}) \\
  = &~ - \mathbb{E}_{(x_H,y_H) \sim (X_H,Y_H)}\sum_{k=1}^{K}\hat{y}_k^H\log(\tilde{y}_k^H)\\
  &~ - \mathbb{E}_{(x_L,y_L) \sim (X_L,Y_L)}\sum_{k=1}^{K}\hat{y}_k^L\log(\tilde{y}_k^L),
  \end{aligned}
  \label{eq:cls}
\end{equation}
where $K$ is the number of identities (classes). We note that weighted classification loss~\cite{chen2017deep} can also be adopted to improve the identity classification performance.

To further enhance the discriminative property, we impose a triplet loss $\mathcal{L}_\mathrm{tri}$ on the feature vector $\mathbfit{v}$, which would maximize the inter-class discrepancy while minimizing intra-class distinctness. To be more specific, for each input image $x$, we sample a positive image $x_\mathrm{pos}$ with the same identity label and a negative image $x_\mathrm{neg}$ with different identity labels to form a triplet tuple. Then, the following equations compute the distances between $x$ and $x_\mathrm{pos}$/$x_\mathrm{neg}$:
\begin{equation}
  \begin{aligned}
  d_\mathrm{pos} = \|\mathbfit{v}_x - \mathbfit{v}_{x_\mathrm{pos}}\|_2,
  \end{aligned}
  \label{eq:d-pos}
\end{equation}
\begin{equation}
  \begin{aligned}
  d_\mathrm{neg} = \|\mathbfit{v}_x - \mathbfit{v}_{x_\mathrm{neg}}\|_2,
  \end{aligned}
  \label{eq:d-neg}
\end{equation}
where $\mathbfit{v}_x$, $\mathbfit{v}_{x_\mathrm{pos}}$, and $\mathbfit{v}_{x_\mathrm{neg}}$ represent the feature vectors of images $x$, $x_\mathrm{pos}$, and $x_\mathrm{neg}$, respectively.

With the above definitions, we have the triplet loss $\mathcal{L}_\mathrm{tri}$ defined as
\begin{equation}
  \begin{aligned}
  \mathcal{L}_\mathrm{tri}&(X_H, X_L; \mathcal{F}, \mathcal{C}) \\
  = &~ \mathbb{E}_{(x_H,y_H) \sim (X_H,Y_H)}\max(0, m + d_\mathrm{pos}^H - d_\mathrm{neg}^H) \\
  + &~ \mathbb{E}_{(x_L,y_L) \sim (X_L,Y_L)}\max(0, m + d_\mathrm{pos}^L - d_\mathrm{neg}^L),
  \end{aligned}
  \label{eq:tri}
\end{equation}
where $m > 0$ is the margin used to define the distance difference between the distance of positive image pair $d_\mathrm{pos}$ and the distance of negative image pair $d_\mathrm{neg}$.

We note that minimizing the triplet loss in~\eqref{eq:tri} is equivalent to minimizing the intra-class distinctness in~\eqref{eq:d-pos} while maximizing the inter-class discrepancy in~\eqref{eq:d-neg}.

{\flushleft {\bf Total loss.}}
Finally, the total loss function $\mathcal{L}$ for training the proposed RAIN is summarized as follows:
\begin{equation}
  \begin{aligned}
  \mathcal{L}&(X_H, X_L; \mathcal{F}, \mathcal{G}, \mathcal{D}_j, \mathcal{C}) \\
  & = \mathcal{L}_\mathrm{adv}^{D_j}(X_H, X_L; \mathcal{F}, \mathcal{D}_j) + \mathcal{L}_\mathrm{rec}(X_H, X_L; \mathcal{F}, \mathcal{G}) \\
  & + \mathcal{L}_\mathrm{cls}(X_H, X_L; \mathcal{F}, \mathcal{C}) + \mathcal{L}_\mathrm{tri}(X_H, X_L; \mathcal{F}, \mathcal{C}).
  \end{aligned}
  \label{eq:fullobj}
\end{equation}

With the above total loss, we aim to solve the min-max criterion:
\begin{equation}
  \min_{\mathcal{F}, \mathcal{G}, \mathcal{C}}\max_{\mathcal{D}_j}\mathcal{L}(X_H, X_L; \mathcal{F}, \mathcal{G}, \mathcal{D}_j.\mathcal{C}).
  \label{eq:min-max-fullobj}
\end{equation}

In other words, to train our RAIN using training HR images (and the down-sampled LR ones), we suppress the classification loss $\mathcal{L}_\mathrm{cls}$, the triplet loss $\mathcal{L}_\mathrm{tri}$, and the HR reconstruction loss $\mathcal{L}_\mathrm{rec}$ while matching feature representations across resolutions.

\begin{table*}[!htbp]
  \small
  \ra{1.3}
  \begin{center}
  \caption{Experimental results of cross-resolution person re-ID (\%). Note that the numbers in bold denote the best results.}
  \label{table:exp-ReID}
  \resizebox{\linewidth}{!} 
  {
  \begin{tabular}{l|cccc|cccc|cccc}
  \toprule
  \multirow{2}{*}{Method} & \multicolumn{4}{c|}{MLR-CUHK03} & \multicolumn{4}{c|}{MLR-VIPeR} & \multicolumn{4}{c}{CAVIAR} \\
  & Rank 1 & Rank 5 & Rank 10 & Rank 20 & Rank 1 & Rank 5 & Rank 10 & Rank 20 & Rank 1 & Rank 5 & Rank 10 & Rank 20 \\
  \midrule
  JUDEA~\cite{li2015multi} & 26.2 & 58.0 & 73.4 & 87.0 & 26.0 & 55.1 & 69.2 & 82.3 & 22.0 & 60.1 & 80.8 & 98.1\\
  SLD$^2$L~\cite{jing2015super} & - & - & - & - & 20.3 & 44.0 & 62.0 & 78.2 & 18.4 & 44.8 & 61.2 & 83.6 \\
  SDF~\cite{wang2016scale} & 22.2 & 48.0 & 64.0 & 80.0 & 9.25 & 38.1 & 52.4 & 68.0 & 14.3 & 37.5 & 62.5 & 95.2 \\
  SING~\cite{jiao2018deep} & 67.7 & 90.7 & 94.7 & 97.4 & 33.5 & 57.0 & 66.5 & 76.6 & 33.5 & 72.7 & 89.0 & 98.6 \\
  CSR-GAN~\cite{wang2018cascaded} & - & - & - & - & 37.2 & 62.3 & 71.6 & 83.7 & - & - & - & - \\
  \midrule
  Baseline (train on HR) & 60.6 & 89.4 & 95.0 & 98.4 & 32.5 & 59.2 & 69.0 & 76.2 & 27.5 & 63.2 & 79.3 & 92.2 \\
  Baseline (train on HR \& LR) & 65.9 & 92.1 & 97.4 & 98.9 & 36.6 & 62.3 & 70.9 & 82.2 & 31.7 & 68.4 & 84.2 & 94.1 \\
  \midrule
  Ours (single-level) & 77.6 & 96.2 & 98.5 & 99.3 & 41.2 & 66.3 & 75.6 & 87.1 & 41.5 & 75.3 & 85.6 & 95.8 \\
  Ours (multi-level)  & \textbf{78.9} & \textbf{97.3} & \textbf{98.7} & \textbf{99.5} & \textbf{42.5} & \textbf{68.3} & \textbf{79.6} & \textbf{88.0} & \textbf{42.0} & \textbf{77.3} & \textbf{89.6} & \textbf{98.7} \\
  \bottomrule
  \end{tabular}
  }
  \end{center}
\end{table*}

\section{Experiments}
We describe the datasets and settings for evaluation. 
%
%Implementation details and additional results available at \href{https://yunchunchen.github.io/RAIN/}{https://yunchunchen.github.io/RAIN/}.

\subsection{Datasets}
We perform evaluations on three benchmark datasets, including two synthetic and one real-world person re-ID datasets. We will explain how we synthesize the LR images for each dataset to perform multiple low-resolution (MLR) person re-ID.

{\flushleft {\bf MLR-CUHK03.}}
The MLR-CUHK03 dataset is a synthetic dataset built from CUHK03~\cite{li2014deepreid} which consists of $5$ different camera views with more than $14,000$ images of $1,467$ person identities. For each camera pair, we down-sample images of one camera by randomly selecting a down-sampling rate $r \in \{2, 3, 4\}$ (\ie the size of the down-sampled image will be $\frac{H}{r} \times \frac{W}{r} \times 3$), while the image resolution of the other camera view remains the same.

{\flushleft {\bf MLR-VIPeR.}}
The MLR-VIPeR dataset is a synthetic dataset built from VIPeR~\cite{gray2008viewpoint} which contains $632$ person-image pairs captured by two cameras. Similarly, we down-sample all the images captured by one camera view using the randomly selected down-sampling rate $r \in \{2, 3, 4\}$, while the image resolution of the other camera is fixed.

{\flushleft {\bf CAVIAR.}}
The more challenging CAVIAR dataset~\cite{Cheng:BMVC11} is a genuine LR person re-ID dataset which contains $1,220$ images of $72$ person identities captured from two camera views. Since the images captured by the more distant camera have much lower resolution than those captured by the closer camera, this dataset is suitable for evaluating the cross-resolution person re-ID. Following \cite{jiao2018deep}, we discard $22$ people who only appear in the closer camera. In contrast to other synthetic datasets, this dataset inherently contains multiple realistic resolutions.

\subsection{Experimental Settings and Protocols}
We consider cross-resolution person re-ID where the query set contains LR images while the gallery set is composed of HR images only. We adopt the standard single-shot person re-ID setting in all of our experiments. Following \cite{wang2016scale}, we randomly divide the MLR-VIPeR and the CAVIAR datasets into halves for training and test set, with $1,367$/$100$ training/test identity split for the MLR-CUHK03 dataset. The test (query) set is constructed with all LR images for each person identity while the gallery image set contains one randomly selected HR image for each person.

For performance evaluation, we adopt the average cumulative match characteristic and report the results recorded at ranks $1$, $5$, $10$, and $20$. We adopt the multi-level discriminator which adapts feature distributions at different feature levels. Due to the balance between efficiency and performance, we select the index of feature level with $j \in \{4, 5\}$ and denote our method as ``Ours (multi-level)'' and the variant of our method with single-level discriminator ($j = 5$) as ``Ours (single-level)''.

\subsection{Evaluation and Comparisons}
We compare our approach with the JUDEA~\cite{li2015multi}, the SLD$^2$L~\cite{jing2015super}, the SDF~\cite{wang2016scale}, the SING~\cite{jiao2018deep}, and the CSR-GAN~\cite{wang2018cascaded}. We note that our cross-resolution feature extractor is only pre-trained on the ImageNet~\cite{he2016deep}. However, SING~\cite{jiao2018deep} and CSR-GAN~\cite{wang2018cascaded} require their re-ID networks to be pre-trained on large-scale re-ID datasets like Market1501~\cite{zheng2015scalable} (which contains $32,668$ images of $1,501$ person identities).

Table~\ref{table:exp-ReID} lists the quantitative results on the three datasets.
We note that our results can be further improved by applying pre-processing or post-processing method, attention mechanisms, or re-ranking. For fair comparisons, no such techniques are applied.

{\flushleft {\bf MLR-CUHK03.}}
Our method achieves $77.6$\% for single-level discriminator and $78.9$\% for multi-level discriminator at rank $1$. The proposed method performs favorably against the state-of-the-art methods and outperforms the previous best competitor~\cite{jiao2018deep} by a large margin $9.9$\% for single-level discriminator and $11.2$\% for multi-level discriminator at rank $1$. Our performance gains can be ascribed to the following two factors. First, unlike most existing re-ID methods, our model performs cross-resolution person re-ID and is trained in an end-to-end learning fashion. Second, the proposed approach would not suffer from the visual artifacts as our model does not leverage SR models.

Furthermore, the advantage of training on both HR and LR images, and introducing the discriminator can be observed by comparing the method ``Ours (single-level)'' with two baseline methods ``Baseline (train on HR)'' and ``Baseline (train on HR \& LR)'', respectively. Note that the method ``Baseline (train on HR)'' is considered as a naive method that only trains on HR images. A $17.0$\% performance drop can be observed from the method ``Baseline (train on HR)''. This indicates that the resolution mismatch problem significantly alters the performance if the model is trained on HR images only. On the other hand, the method ``Baseline (train on HR \& LR)'' trained on both HR and LR images without applying the adversarial loss still suffers a $11.7$\% performance drop. The result suggests that even if the model is trained with images of multiple resolutions, without the adversarial loss which aligns image features across resolutions, the resolution mismatch problem still implicitly alters the performance.

%% Comparison table
\begin{table}[t]
  \scriptsize
  \caption{Experimental results of cross-resolution person re-ID with seen and unseen resolutions on the MLR-VIPeR dataset evaluated at rank 1 (\%).}
  \centering
  \label{table:comparison}
  \resizebox{\linewidth}{!}
  {
  \begin{tabular}{l|l|ccc}
  \toprule
  \multirow{2}{*}{Train} & \multirow{2}{*}{Test} & \multicolumn{3}{c}{MLR-VIPeR} \\
  & & SING & CSR-GAN & Ours \\
  \midrule
  $r \in \{1, 2, 3, 4\}$ & $r \in \{2, 3, 4\}$ & 33.5 & 37.2 & 42.5 \\
  $r \in \{1, 2, 3, 4\}$  & $r = 8$ & \xmark & \xmark & 37.5 \\
  \bottomrule
  \end{tabular}
  }
\end{table}

%% t-SNE Visualization
\setlength{\twoimg}{1.0\linewidth}
\begin{figure*}[t]
  \centering
  \begin{subfigure}[b]{\twoimg}
    \centering\includegraphics[width=\linewidth]{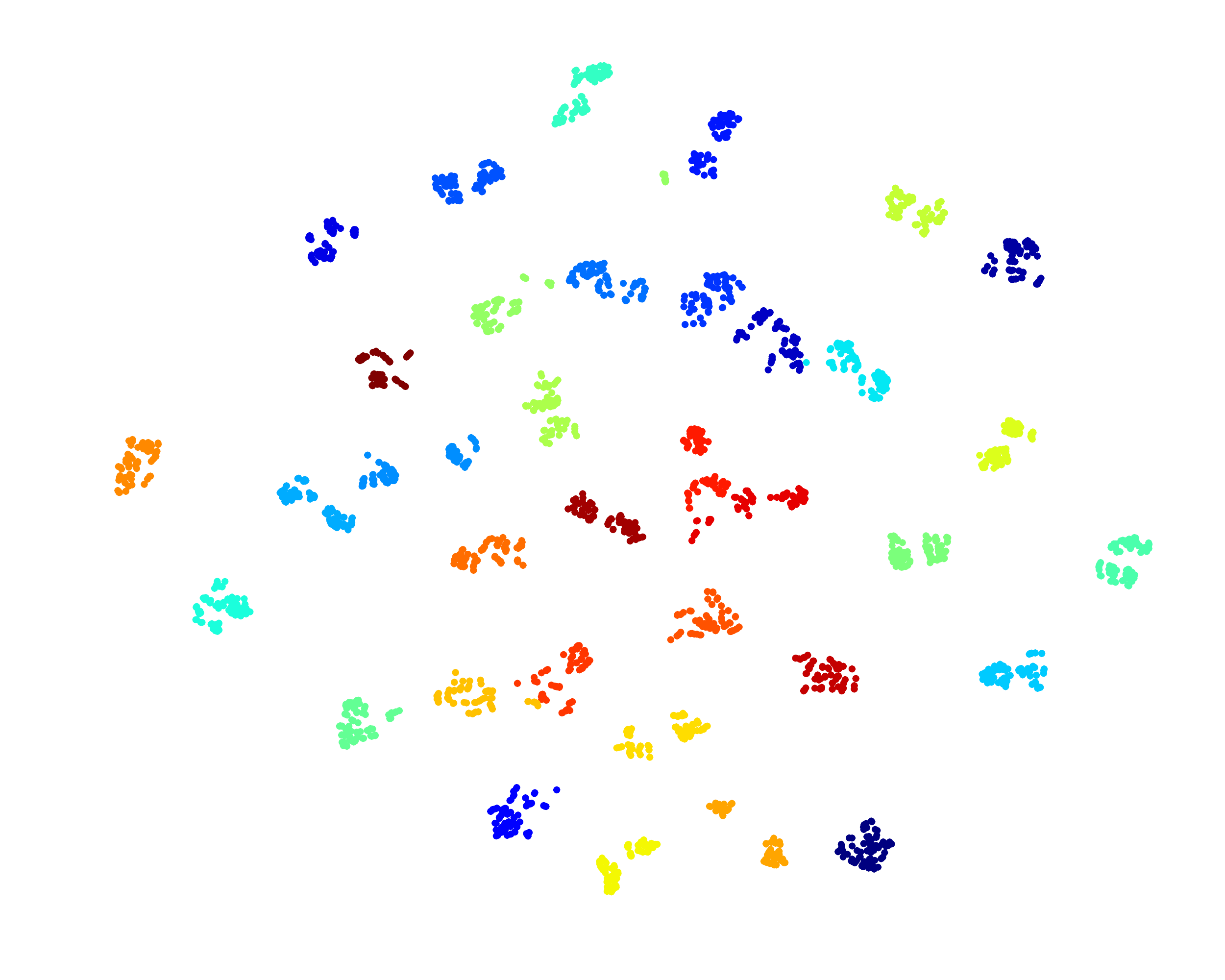}\\
    \caption{Colorization with respect to \textbf{identity}.}
    \label{fig:tsne-identity}
  \end{subfigure}
  \begin{subfigure}[b]{\twoimg}
    \centering\includegraphics[width=\linewidth]{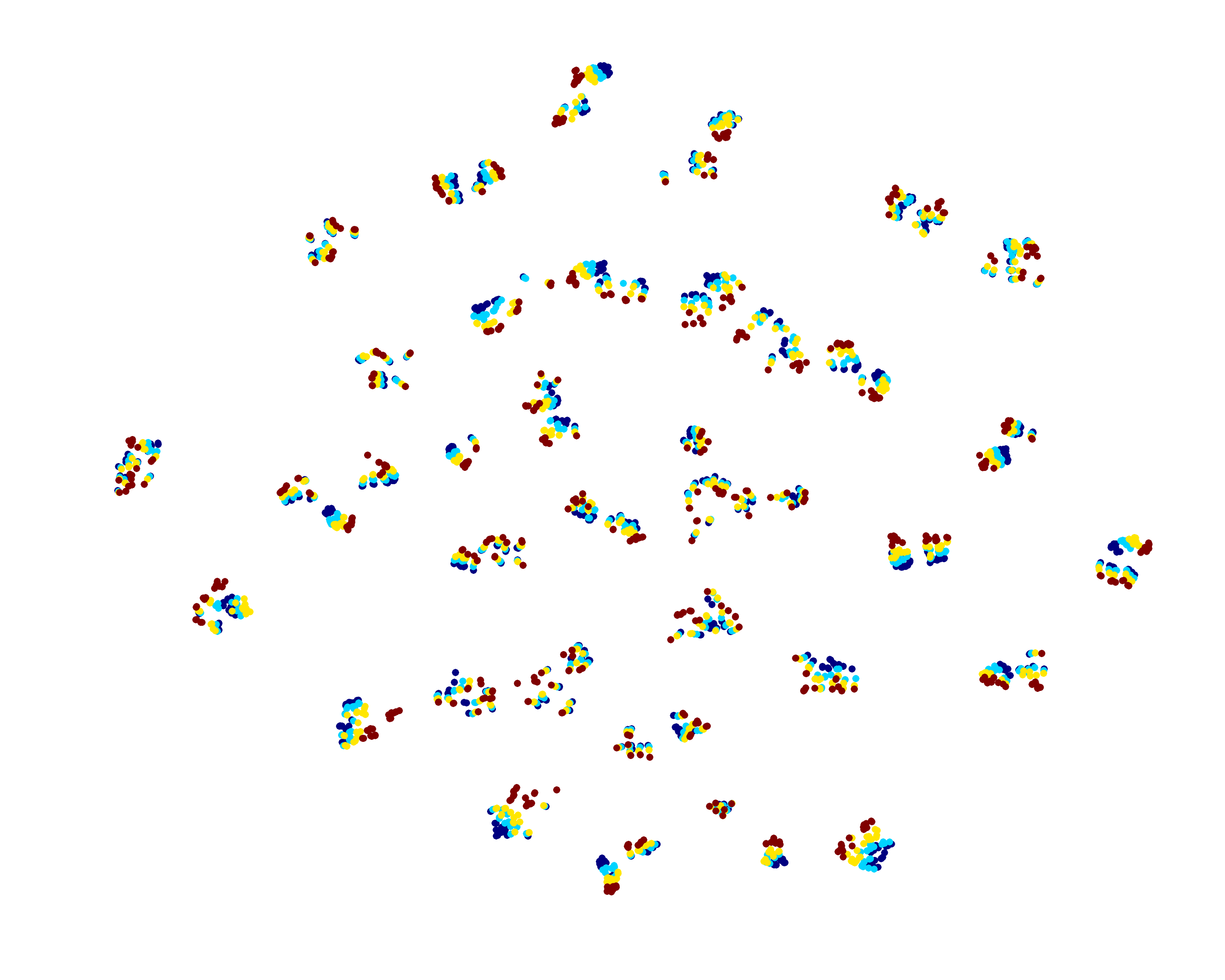}\\
    \caption{Colorization with respect to \textbf{resolution}.}
    \label{fig:tsne-resolution}
  \end{subfigure}
  \caption{Visualization of cross-resolution feature vectors $\mathbfit{v}$ on MLR-CUHK03 via t-SNE.
  (a) $35$ different identities, each of which is shown in a unique color.
  (b) With four different down-sampling rates ($r \in \{1, 2, 4, 8\}$) are considered and shown, images with the same resolution are shown in the same color. Note that images with $r = 8$ are not seen during training.
  }
\end{figure*}

{\flushleft {\bf MLR-VIPeR.}}
Our method achieves the state-of-the-art performance on all four ranks. The performance gains over the best competitor~\cite{wang2018cascaded} at rank $1$ are $4.0$\% for single-level discriminator and $5.3$\% for multi-level discriminator.

In addition to performance comparisons, our method can reliably perform cross-resolution person re-ID and generalize well on unseen image resolutions. However, most existing methods~\cite{jiao2018deep,wang2018cascaded} may not properly handle image resolutions that are not seen by their SR models or require to fuse the results produced by multiple learned models, each of which is specifically designed for a particular resolution.

We present and compare such results in Table~\ref{table:comparison}. Suppose that the training set contains images with different down-sampling rates $r \in \{1, 2, 3, 4\}$ ($r = 1$ indicates that images remain their original sizes), if the test set contains images with down-sampling rates $r \in \{2, 3, 4\}$ which have appeared in the training set, both existing methods~\cite{jiao2018deep,wang2018cascaded} and our approach perform cross-resolution person re-ID properly. However, if we consider another scenario where the training set contains images with down-sampling rates $r \in \{1, 2, 3, 4\}$, whereas the test set contains images with down-sampling rate $r = 8$ which are not seen during training, the proposed model works properly and reliably performs cross-resolution person re-ID with satisfactory result. However, existing methods could not handle unseen resolutions properly due to the following reasons. For CSR-GAN~\cite{wang2018cascaded}, their SR models are resolution-dependent and cannot directly apply to LR images of unseen resolutions for synthesizing HR images. For SING~\cite{jiao2018deep}, even though they can still apply their SR models on the images of unseen resolutions via fusing the results produced by several models, their SR models are specifically designed for some particular image resolutions, which will not reliably address cross-resolution person re-ID with images of unseen resolutions.

{\flushleft {\bf CAVIAR.}}
For the CAVIAR dataset, our method achieves $41.5$\% for single-level discriminator and $42.0$\% for multi-level discriminator at rank $1$ score achieving the state-of-the-art performance on all four evaluated ranks. The performance gains over the best competitor~\cite{jiao2018deep} measured at rank $1$ are $8.0$\% for single-level discriminator and $8.5$\% for multi-level discriminator.

%% Ablation study
\begin{table}[t]
  \small
  \ra{1.3}
  \begin{center}
  \caption{Ablation studies on the MLR-CUHK03 dataset (\%).}
  \label{table:exp-ablation}
  \resizebox{\linewidth}{!} 
  {
  \begin{tabular}{l|ccccc}
  \toprule
  \multirow{2}{*}{Method} & \multicolumn{5}{c}{MLR-CUHK03} \\
  & Rank 1 & Rank 5 & Rank 10 & Rank 20 & mAP \\
  \midrule
  Ours & \textbf{78.9} & \textbf{97.3} & \textbf{98.7} & \textbf{99.5} & \textbf{74.5} \\
  Ours w/o $\mathcal{L}_\mathrm{cls}$ & 70.8 & 95.1 & 97.7 & 98.9 & 68.0 \\
  Ours w/o $\mathcal{L}_\mathrm{tri}$ & 69.1 & 92.2 & 96.6 & 98.7 & 64.1 \\
  Ours w/o $\mathcal{L}_\mathrm{rec}$ & 67.3 & 89.5 & 94.5 & 97.7 & 64.2\\
  Ours w/o $\mathcal{L}_\mathrm{adv}$ & 65.9 & 92.1 & 97.4 & 98.9 & 62.3\\
  \bottomrule
  \end{tabular}
  }
  \end{center}
\end{table}

\begin{table}[t]
  \small
  \ra{1.3}
  \begin{center}
  \caption{Effect of training images of multiple low resolutions. The bold numbers indicate the best results (\%).}
  \label{table:exp-multi-resolution}
  \resizebox{\linewidth}{!} 
  {
  \begin{tabular}{ll|cc|cc|cc}
  \toprule
  \multicolumn{2}{l|}{Train} & \multicolumn{2}{c|}{MLR-CUHK03} & \multicolumn{2}{c|}{MLR-VIPeR} & \multicolumn{2}{c}{CAVIAR} \\
  HR & LR & Rank 1 & mAP & Rank 1 & mAP & Rank 1 & mAP \\
  \midrule
  $r = 1$ & $r = 2$ & 70.9 & 67.7 & 38.9 & 42.6 & 36.3 & 52.9\\
  $r = 1$ & $r = 3$ & 72.3 & 68.8 & 40.3 & 43.2 & 37.9 & 53.1\\
  $r = 1$ & $r = 4$ & 77.2 &74.8 & 41.5 & 45.4 & 40.1 & 54.6\\
  $r = 1$ & $r \in \{2, 3, 4\}$ & \textbf{78.9} & \textbf{75.9} & \textbf{42.5}& \textbf{47.0}& \textbf{42.0}& \textbf{56.3}\\
  \bottomrule
  \end{tabular}
  }
  \end{center}
\end{table}

\subsection{Ablation Studies}

{\flushleft {\bf Loss functions.}} To analyze the importance of each loss function, we conduct an ablation study on the MLR-CUHK03 dataset using the multi-level discriminator method abbreviated as ``Ours''. Table~\ref{table:exp-ablation} presents the quantitative results of the ablation experiments evaluated at ranks $1$, $5$, $10$, and $20$, and the mAP. The results show that without the classification loss $\mathcal{L}_\mathrm{cls}$ or the triplet loss $\mathcal{L}_\mathrm{tri}$, our model still achieves favorable performances compared with the state-of-the-art method~\cite{jiao2018deep}. This is because both the classification loss and the triplet loss are introduced to control the intra-class and inter-class distances. Without either one of them, our model still has the ability to establish a well separated feature space for each person identity. However, our model suffers a $11.6$\% performance drop without the HR reconstruction loss $\mathcal{L}_\mathrm{rec}$ at rank $1$. The result indicates that introducing the HR decoder and imposing the HR reconstruction loss greatly reduce the information loss. In addition, without the adversarial loss $\mathcal{L}_\mathrm{adv}$, our model does not learn resolution-invariant representations and a $13.0$\% performance drop at rank $1$ can be observed, which indicates that our model suffers from the severe impact induced by the resolution mismatch problem. The ablation experiments show that all loss terms play crucial roles in achieving state-of-the-art performance.

\begin{figure}[t]
  \centering
  \includegraphics[width=\linewidth]{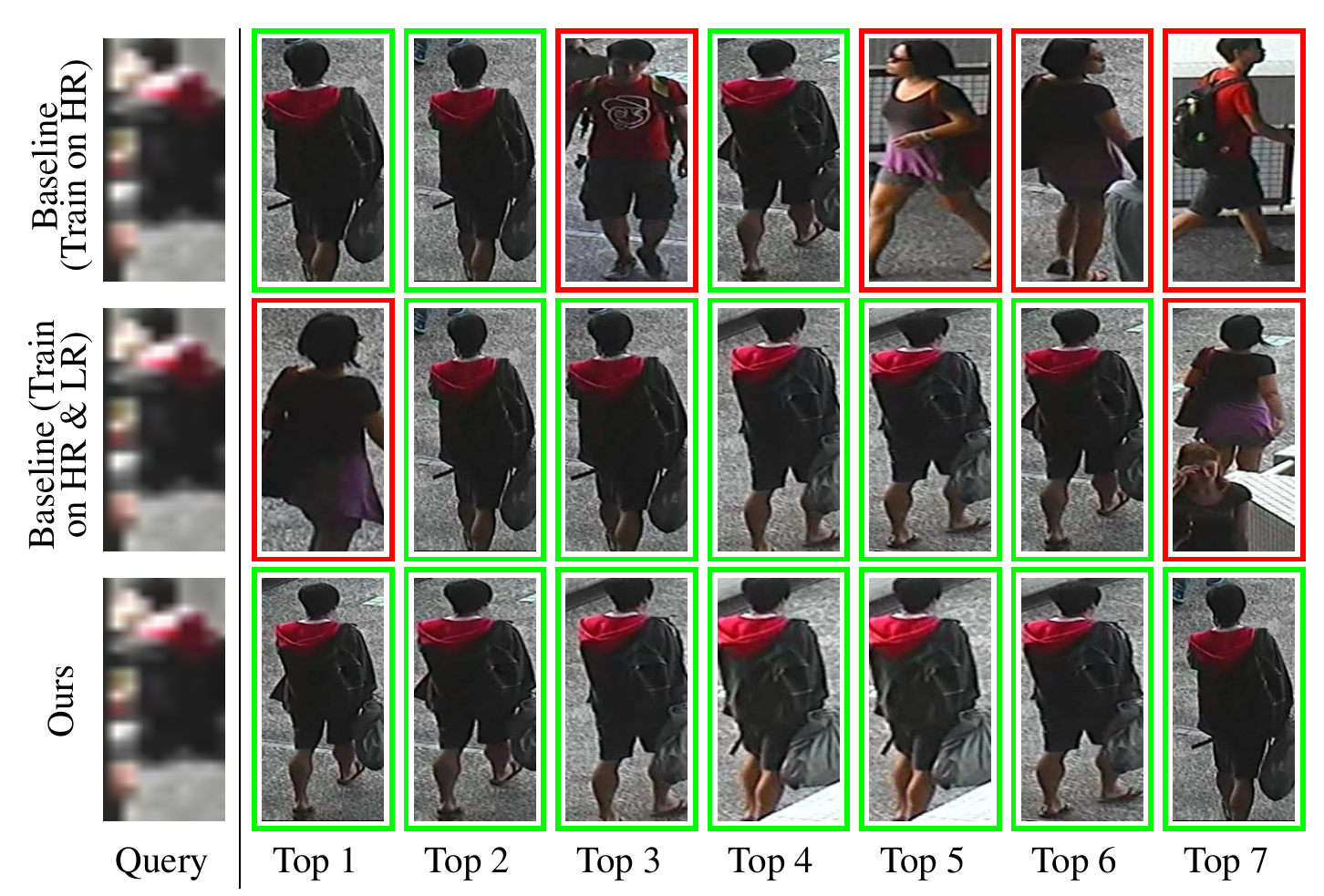}
  \caption{Example of the top-ranked HR gallery images of the MLR-CUHK03 dataset which are matched by the LR query input. Images bounded in green and red rectangles denote correct and incorrect matches, respectively.}
  \label{fig:rank}
\end{figure}

{\flushleft {\bf Effect of training images of multiple low resolutions.}} We conduct experiments on all three adopted datasets with different combinations of down-sampling rates and present the results in Table~\ref{table:exp-multi-resolution}. We observe that when our model (bottom row) is trained with LR images of multiple down-sampling rates, it achieves the best results compared with our three variant methods (first three rows), each of which is trained with LR images of a single down-sampling rate. On the other hand, from the third variant method (the third row), the result demonstrates that our method reliably handles unseen but intermediate resolutions ($r \in \{2, 3\}$).

{\flushleft {\bf Visualization of cross-resolution feature vector $\mathbfit{v}$.}}
We now visualize the feature vectors $\mathbfit{v}$ on the test set of the MLR-CUHK03 dataset in Figures~\ref{fig:tsne-identity} and~\ref{fig:tsne-resolution} via t-SNE.

In Figure~\ref{fig:tsne-identity}, we select $35$ different person identities, each of which is indicated by a color. We observe that the projected feature vectors are well separated, which suggests that sufficient re-ID ability can be exhibited by our model. On the other hand, for Figure~\ref{fig:tsne-resolution}, we colorize each image resolution with a color in each identity cluster (four different down-sampling rates $r \in \{1, 2, 4, 8\}$). It can be observed that the projected feature vectors of the same identity but different down-sampling rates are all well clustered. We note that images with down-sampling rate $r = 8$ are not presented in the training set.

The above visualizations demonstrate that our model learns resolution-invariant representations, and is able to generalize well to unseen image resolution (\eg $r = 8$) for cross-resolution person re-ID.

{\flushleft {\bf Top ranked gallery images.}}
Given an LR query image with down-sampling rate $r = 8$ (the leftmost column), we present the first $7$ top-ranked HR gallery images in Figure~\ref{fig:rank}. We compare our method (bottom row) with two baseline methods ``Baseline (Train on HR)'' (top row) and ``Baseline (Train on HR and LR)'' (middle row). The green and red boundaries indicate correct and incorrect matches, respectively. From the results in the top row of this figure, we observe that the method ``Baseline (Train on HR)'' only achieves $3$ out of $7$ correct matches. When trained with images of various resolutions, the method ``Baseline (Train on HR and LR)'' improves the matching results to $5$ out of $7$ correct matches. Finally, our method achieves $7$ out of $7$ correct matches, which again verify the effectiveness and robustness of our model. Note that the resolution ($r = 8$) of the query image is not seen during training.

\begin{figure}[t]
  \centering
  \includegraphics[width=\linewidth]{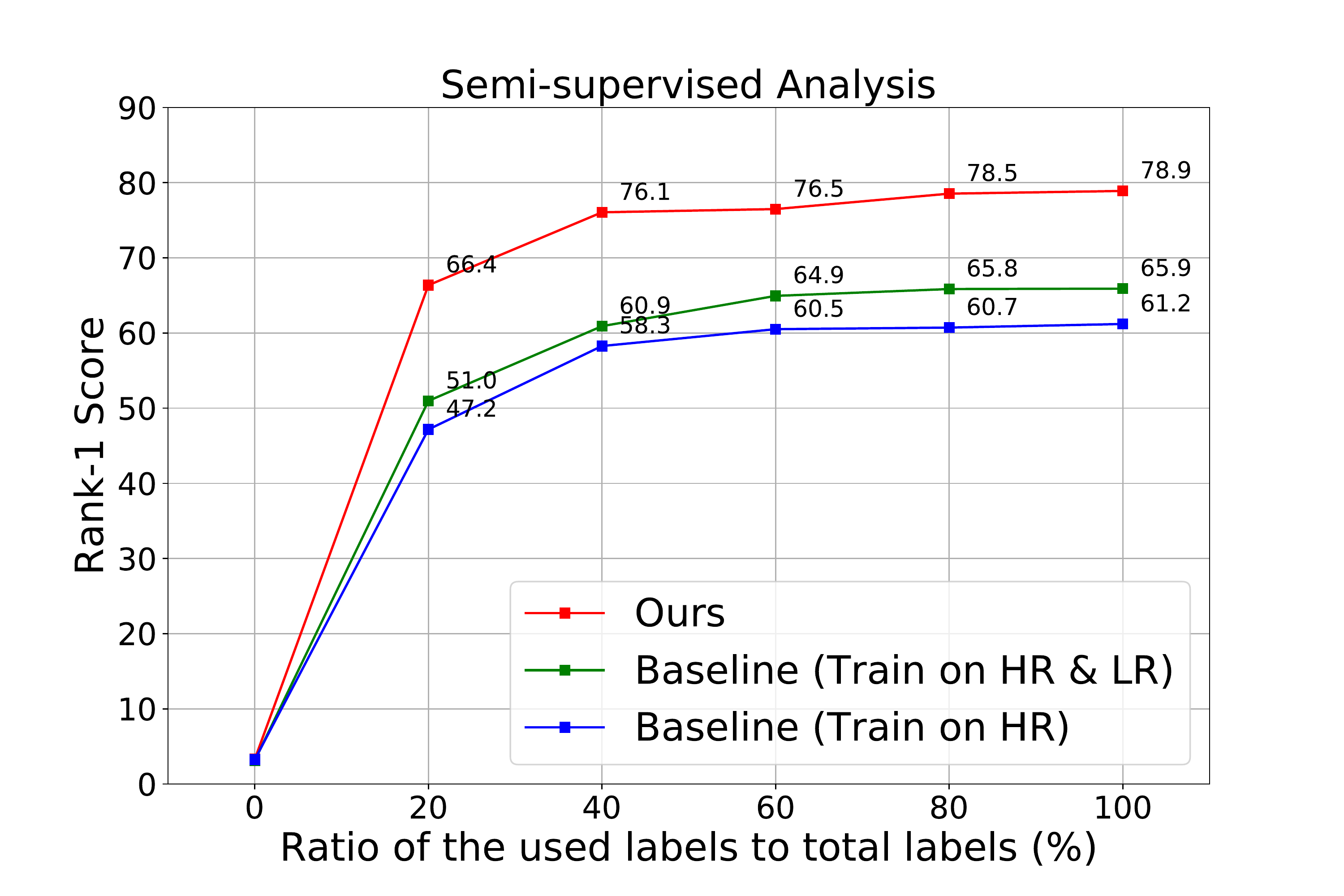}
  \caption{Semi-supervised cross-resolution person re-ID on the MLR-CUHK03 dataset (\%).}
  \label{fig:exp-semi}
\end{figure}

\subsection{Semi-Supervised Cross-Resolution Re-ID}

To show that, even only a portion of the dataset are with labels (\ie computing the classification loss and the triplet loss is only applicable for such data), the unique design of our RAIN would still exhibit sufficient ability in learning cross-resolution person re-ID image features, we conduct a series of semi-supervised experiments.

We increase the amount of labeled data by $20$\% each time (\ie $0$\%, $20$\%, $40$\%, $60$\%, $80$\%, and $100$\% labeled data) and record the performance at rank $1$ as presented in Figure~\ref{fig:exp-semi}. Note that the unlabeled data can still compute the HR reconstruction loss and the adversarial loss. We compare our method with two baseline methods ``Baseline (train on HR)'' and ``Baseline (train on HR \& LR)''. From Figure~\ref{fig:exp-semi}, we observe that without any labeled information, our method achieves $3.3$\% at rank $1$. When the amount of labeled data is increased to $20$\%, our model results in $66.3$\% at rank $1$, which is only slightly worse than the performance of SING~\cite{jiao2018deep} ($67.7$\%) recorded with $100$\% labeled data. When we further increase the amount of labeled data to $40$\%, our model achieves $76.1$\% at rank $1$, which outperforms the result of SING~\cite{jiao2018deep} recorded with fully labeled data by $8.4$\%.

With the experiments, we confirm that the unique design and the integration of cross-resolution feature extractor (with resolution adversarial learning), HR decoder, and classification components, would allow one to learn resolution-invariant representations for cross-resolution person re-ID, even if only a portion of the image data are labeled. Thus, the use of our RAIN for real-world re-ID problems can be supported.

\section{Conclusions}
We have presented an \emph{end-to-end trainable} network, Resolution Adaptation and re-Identification Network (RAIN), which is able to learn \emph{resolution-invariant} representations for cross-resolution person re-ID. The novelty of this network lies in the use of adversarial learning for deriving latent image features across image resolutions, with an autoencoder-like architecture which preserves the image representation ability. Utilizing image labels, the classification components further exploit the discriminative property for re-ID purposes. From our experiments, we confirm that our model performs favorably against state-of-the-art cross-resolution person re-ID methods. We also verify that our model is able to handle LR query inputs with varying image resolutions, even if such resolutions are not seen during training. Finally, the extension to semi-supervised re-ID further supports the use of our proposed model for solving practical cross-resolution re-ID tasks. In the future, we hope to leverage attention mechanisms by localizing~\cite{SaliencyAware} or segmenting~\cite{VIPCup} identities to better focus on classifying foreground objects. We also hope our proposed method can facilitate other computer vision tasks such as semantic matching~\cite{WeakMatchNet}, object co-segmentation~\cite{MaCoSNet}, and domain adaptation~\cite{CrDoCo}.
\\\\
\noindent\textbf{Acknowledgements.}
This work is supported in part by Umbo Computer Vision and the Ministry of Science and Technology of Taiwan under grant MOST 107-2634-F-002-010.
\bibliographystyle{aaai}
\bibliography{reference}

\begin{thebibliography}{}

\bibitem[\protect\citeauthoryear{Chang, Hospedales, and
  Xiang}{2018}]{chang2018multi}
Chang, X.; Hospedales, T.~M.; and Xiang, T.
\newblock 2018.
\newblock Multi-level factorisation net for person re-identification.
\newblock In {\em CVPR}.

\bibitem[\protect\citeauthoryear{Chen and Hsu}{2019}]{SaliencyAware}
Chen, Y.-C., and Hsu, W.~H.
\newblock 2019.
\newblock Saliency aware: Weakly supervised object localization.
\newblock In {\em ICASSP}.

\bibitem[\protect\citeauthoryear{Chen \bgroup et al\mbox.\egroup
  }{2017}]{chen2017deep}
Chen, Y.-C.; Li, Y.-J.; Tseng, A.; and Lin, T.
\newblock 2017.
\newblock Deep learning for malicious flow detection.
\newblock In {\em IEEE PIMRC}.

\bibitem[\protect\citeauthoryear{Chen \bgroup et al\mbox.\egroup
  }{2018a}]{chen2018group}
Chen, D.; Xu, D.; Li, H.; Sebe, N.; and Wang, X.
\newblock 2018a.
\newblock Group consistent similarity learning via deep crf for person
  re-identification.
\newblock In {\em CVPR}.

\bibitem[\protect\citeauthoryear{Chen \bgroup et al\mbox.\egroup
  }{2018b}]{WeakMatchNet}
Chen, Y.-C.; Huang, P.-H.; Yu, L.-Y.; Huang, J.-B.; Yang, M.-H.; and Lin, Y.-Y.
\newblock 2018b.
\newblock Deep semantic matching with foreground detection and cycle
  consistency.
\newblock In {\em ACCV}.

\bibitem[\protect\citeauthoryear{Chen \bgroup et al\mbox.\egroup
  }{2019a}]{CrDoCo}
Chen, Y.-C.; Lin, Y.-Y.; Yang, M.-H.; and Huang, J.-B.
\newblock 2019a.
\newblock Crdoco: Pixel-level domain transfer with cross-domain consistency.
\newblock In {\em CVPR}.

\bibitem[\protect\citeauthoryear{Chen \bgroup et al\mbox.\egroup
  }{2019b}]{MaCoSNet}
Chen, Y.-C.; Lin, Y.-Y.; Yang, M.-H.; and Huang, J.-B.
\newblock 2019b.
\newblock Show, match and segment: Joint learning of semantic matching and
  object co-segmentation.
\newblock {\em arXiv}.

\bibitem[\protect\citeauthoryear{Cheng \bgroup et al\mbox.\egroup
  }{2011}]{Cheng:BMVC11}
Cheng, D.~S.; Cristani, M.; Stoppa, M.; Bazzani, L.; and Murino, V.
\newblock 2011.
\newblock Custom pictorial structures for re-identification.
\newblock In {\em BMVC}.

\bibitem[\protect\citeauthoryear{Cheng \bgroup et al\mbox.\egroup
  }{2016}]{cheng2016person}
Cheng, D.; Gong, Y.; Zhou, S.; Wang, J.; and Zheng, N.
\newblock 2016.
\newblock Person re-identification by multi-channel parts-based cnn with
  improved triplet loss function.
\newblock In {\em CVPR}.

\bibitem[\protect\citeauthoryear{Goodfellow \bgroup et al\mbox.\egroup
  }{2014}]{goodfellow2014generative}
Goodfellow, I.; Pouget-Abadie, J.; Mirza, M.; Xu, B.; Warde-Farley, D.; Ozair,
  S.; Courville, A.; and Bengio, Y.
\newblock 2014.
\newblock Generative adversarial nets.
\newblock In {\em NIPS}.

\bibitem[\protect\citeauthoryear{Gray and Tao}{2008}]{gray2008viewpoint}
Gray, D., and Tao, H.
\newblock 2008.
\newblock Viewpoint invariant pedestrian recognition with an ensemble of
  localized features.
\newblock In {\em ECCV}.

\bibitem[\protect\citeauthoryear{He \bgroup et al\mbox.\egroup
  }{2016}]{he2016deep}
He, K.; Zhang, X.; Ren, S.; and Sun, J.
\newblock 2016.
\newblock Deep residual learning for image recognition.
\newblock In {\em CVPR}.

\bibitem[\protect\citeauthoryear{Hermans, Beyer, and
  Leibe}{2017}]{hermans2017defense}
Hermans, A.; Beyer, L.; and Leibe, B.
\newblock 2017.
\newblock In defense of the triplet loss for person re-identification.
\newblock {\em arXiv}.

\bibitem[\protect\citeauthoryear{Huang \bgroup et al\mbox.\egroup
  }{2018}]{huang2018munit}
Huang, X.; Liu, M.-Y.; Belongie, S.; and Kautz, J.
\newblock 2018.
\newblock Multimodal unsupervised image-to-image translation.
\newblock In {\em ECCV}.

\bibitem[\protect\citeauthoryear{Jiao \bgroup et al\mbox.\egroup
  }{2018}]{jiao2018deep}
Jiao, J.; Zheng, W.-S.; Wu, A.; Zhu, X.; and Gong, S.
\newblock 2018.
\newblock Deep low-resolution person re-identification.
\newblock In {\em AAAI}.

\bibitem[\protect\citeauthoryear{Jing \bgroup et al\mbox.\egroup
  }{2015}]{jing2015super}
Jing, X.-Y.; Zhu, X.; Wu, F.; You, X.; Liu, Q.; Yue, D.; Hu, R.; and Xu, B.
\newblock 2015.
\newblock Super-resolution person re-identification with semi-coupled low-rank
  discriminant dictionary learning.
\newblock In {\em CVPR}.

\bibitem[\protect\citeauthoryear{Kalayeh \bgroup et al\mbox.\egroup
  }{2018}]{kalayeh2018human}
Kalayeh, M.~M.; Basaran, E.; G{\"o}kmen, M.; Kamasak, M.~E.; and Shah, M.
\newblock 2018.
\newblock Human semantic parsing for person re-identification.
\newblock In {\em CVPR}.

\bibitem[\protect\citeauthoryear{Ledig \bgroup et al\mbox.\egroup
  }{2017}]{ledig2017photo}
Ledig, C.; Theis, L.; Husz{\'a}r, F.; Caballero, J.; Cunningham, A.; Acosta,
  A.; Aitken, A.~P.; Tejani, A.; Totz, J.; Wang, Z.; et~al.
\newblock 2017.
\newblock Photo-realistic single image super-resolution using a generative
  adversarial network.
\newblock In {\em CVPR}.

\bibitem[\protect\citeauthoryear{Li \bgroup et al\mbox.\egroup
  }{2014}]{li2014deepreid}
Li, W.; Zhao, R.; Xiao, T.; and Wang, X.
\newblock 2014.
\newblock Deepreid: Deep filter pairing neural network for person
  re-identification.
\newblock In {\em CVPR}.

\bibitem[\protect\citeauthoryear{Li \bgroup et al\mbox.\egroup
  }{2015}]{li2015multi}
Li, X.; Zheng, W.-S.; Wang, X.; Xiang, T.; and Gong, S.
\newblock 2015.
\newblock Multi-scale learning for low-resolution person re-identification.
\newblock In {\em ICCV}.

\bibitem[\protect\citeauthoryear{Li \bgroup et al\mbox.\egroup }{2019}]{CADNet}
Li, Y.-J.; Chen, Y.-C.; Lin, Y.-Y.; Du, X.; and Wang, Y.-C.~F.
\newblock 2019.
\newblock Recover and identify: A generative dual model for cross-resolution
  person re-identification.
\newblock In {\em ICCV}.

\bibitem[\protect\citeauthoryear{Li, Zhu, and Gong}{2018}]{li2018harmonious}
Li, W.; Zhu, X.; and Gong, S.
\newblock 2018.
\newblock Harmonious attention network for person re-identification.
\newblock In {\em CVPR}.

\bibitem[\protect\citeauthoryear{Lin \bgroup et al\mbox.\egroup
  }{2017}]{lin2017improving}
Lin, Y.; Zheng, L.; Zheng, Z.; Wu, Y.; and Yang, Y.
\newblock 2017.
\newblock Improving person re-identification by attribute and identity
  learning.
\newblock {\em arXiv}.

\bibitem[\protect\citeauthoryear{Lin \bgroup et al\mbox.\egroup
  }{2019}]{VIPCup}
Lin, J.-Y.; Wu, M.-S.; Chang, Y.-C.; Chen, Y.-C.; Chou, C.-T.; Wu, C.-T.; and
  Hsu, W.~H.
\newblock 2019.
\newblock Learning volumetric segmentation for lung tumor.
\newblock {\em IEEE ICIP VIP Cup Tech. report}.

\bibitem[\protect\citeauthoryear{Liu \bgroup et al\mbox.\egroup
  }{2018}]{liu2018pose}
Liu, J.; Ni, B.; Yan, Y.; Zhou, P.; Cheng, S.; and Hu, J.
\newblock 2018.
\newblock Pose transferrable person re-identification.
\newblock In {\em CVPR}.

\bibitem[\protect\citeauthoryear{Shen \bgroup et al\mbox.\egroup
  }{2018}]{shen2018deep}
Shen, Y.; Li, H.; Xiao, T.; Yi, S.; Chen, D.; and Wang, X.
\newblock 2018.
\newblock Deep group-shuffling random walk for person re-identification.
\newblock In {\em CVPR}.

\bibitem[\protect\citeauthoryear{Si \bgroup et al\mbox.\egroup
  }{2018}]{si2018dual}
Si, J.; Zhang, H.; Li, C.-G.; Kuen, J.; Kong, X.; Kot, A.~C.; and Wang, G.
\newblock 2018.
\newblock Dual attention matching network for context-aware feature sequence
  based person re-identification.
\newblock In {\em CVPR}.

\bibitem[\protect\citeauthoryear{Song \bgroup et al\mbox.\egroup
  }{2018}]{song2018mask}
Song, C.; Huang, Y.; Ouyang, W.; and Wang, L.
\newblock 2018.
\newblock Mask-guided contrastive attention model for person re-identification.
\newblock In {\em CVPR}.

\bibitem[\protect\citeauthoryear{Wang \bgroup et al\mbox.\egroup
  }{2016}]{wang2016scale}
Wang, Z.; Hu, R.; Yu, Y.; Jiang, J.; Liang, C.; and Wang, J.
\newblock 2016.
\newblock Scale-adaptive low-resolution person re-identification via learning a
  discriminating surface.
\newblock In {\em IJCAI}.

\bibitem[\protect\citeauthoryear{Wang \bgroup et al\mbox.\egroup
  }{2018a}]{wang2018resource}
Wang, Y.; Wang, L.; You, Y.; Zou, X.; Chen, V.; Li, S.; Huang, G.; Hariharan,
  B.; and Weinberger, K.~Q.
\newblock 2018a.
\newblock Resource aware person re-identification across multiple resolutions.
\newblock In {\em CVPR}.

\bibitem[\protect\citeauthoryear{Wang \bgroup et al\mbox.\egroup
  }{2018b}]{wang2018cascaded}
Wang, Z.; Ye, M.; Yang, F.; Bai, X.; and Satoh, S.
\newblock 2018b.
\newblock Cascaded sr-gan for scale-adaptive low resolution person
  re-identification.
\newblock In {\em IJCAI}.

\bibitem[\protect\citeauthoryear{Wei \bgroup et al\mbox.\egroup
  }{2018}]{wei2018person}
Wei, L.; Zhang, S.; Gao, W.; and Tian, Q.
\newblock 2018.
\newblock Person transfer gan to bridge domain gap for person
  re-identification.
\newblock In {\em CVPR}.

\bibitem[\protect\citeauthoryear{Zheng \bgroup et al\mbox.\egroup
  }{2015}]{zheng2015scalable}
Zheng, L.; Shen, L.; Tian, L.; Wang, S.; Wang, J.; and Tian, Q.
\newblock 2015.
\newblock Scalable person re-identification: A benchmark.
\newblock In {\em ICCV}.

\bibitem[\protect\citeauthoryear{Zheng, Yang, and
  Hauptmann}{2016}]{zheng2016person}
Zheng, L.; Yang, Y.; and Hauptmann, A.~G.
\newblock 2016.
\newblock Person re-identification: Past, present and future.
\newblock {\em arXiv}.

\bibitem[\protect\citeauthoryear{Zhong \bgroup et al\mbox.\egroup
  }{2018}]{zhong2017camera}
Zhong, Z.; Zheng, L.; Zheng, Z.; Li, S.; and Yang, Y.
\newblock 2018.
\newblock Camera style adaptation for person re-identification.
\newblock In {\em CVPR}.

\end{thebibliography}

\end{document}